\documentclass[runningheads]{llncs}

\usepackage{eccv}

\usepackage{eccvabbrv}

\usepackage{graphicx}
\usepackage{booktabs}
\usepackage{pifont}
\usepackage[accsupp]{axessibility}  

\usepackage{hyperref}

\usepackage{orcidlink}
\usepackage[table]{xcolor}
\usepackage{tabularx}
\usepackage{array}
\usepackage{pifont}

\newcolumntype{Y}{>{\centering\arraybackslash}X}

\begin{document}

\title{3rd Place Solution to Human Motion Challenges
in Real-World and Clinical Settings (MoCha) @ECCV2026: Language-Aligned Motion Representations for Domain-Generalizable UPDRS-Gait Severity Estimation}


\author{Soojie Kim \and Muhammad Munsif \and Minkyung Kim \and Seungryul Baek}


\institute{
UNIST, South Korea\\
\email{\{soojie,munsif,minn0219,srbaek\}@unist.ac.kr}}

\maketitle

\begin{abstract}
 In this work, we introduce language-aligned motion representations for domain-generalizable UPDRS-Gait severity estimation, aiming to learn semantically structured motion features that generalize across heterogeneous clinical domains. We first learn motion representations using a Bi-GRU backbone that captures the temporal dynamics of SMPL sequences. Prior to model training, motion captions are generated offline using Qwen2.5-7B-Instruct. The backbone is then trained with both classification and text-alignment objectives to learn discriminative and semantically structured motion representations while accounting for the class imbalance present in the training data. We subsequently adapt the learned backbone independently to each source domain so that the model can capture domain-specific motion characteristics. The resulting source-specific models are then merged at the parameter level to consolidate complementary knowledge across source domains into a single domain-generalized model. To further mitigate class imbalance, we perform GPT-5.5-based pseudo labeling, and our final merged models for each site do not use any class-prior correction during inference. The resulting model is evaluated under the unseen-site setting of the MoCha Challenge, using Macro F1 as the primary evaluation metric. Our method achieves a \textbf{macro-F1 of 0.57} on the hidden test set with only 637K active parameters at inference, \textbf{ranking 3rd among 58 leaderboard entries} in the MoCha 2026 Challenge\footnote{\url{https://mocha.care-pd.ca/challenge.html}.}. The challenge attracted 1,669 submissions from 112 participants and offered monetary prizes sponsored by Machine Medicine Technologies\footnote{\url{https://machinemedicine.io/en/}.}.

 \vspace{0.5em} 

\end{abstract}

\section{Introduction}
\label{sec:intro}

\begin{figure}[htbp]
    \centering
    \includegraphics[width=\linewidth, trim=0cm 1.0cm 0cm 1.0cm, clip]{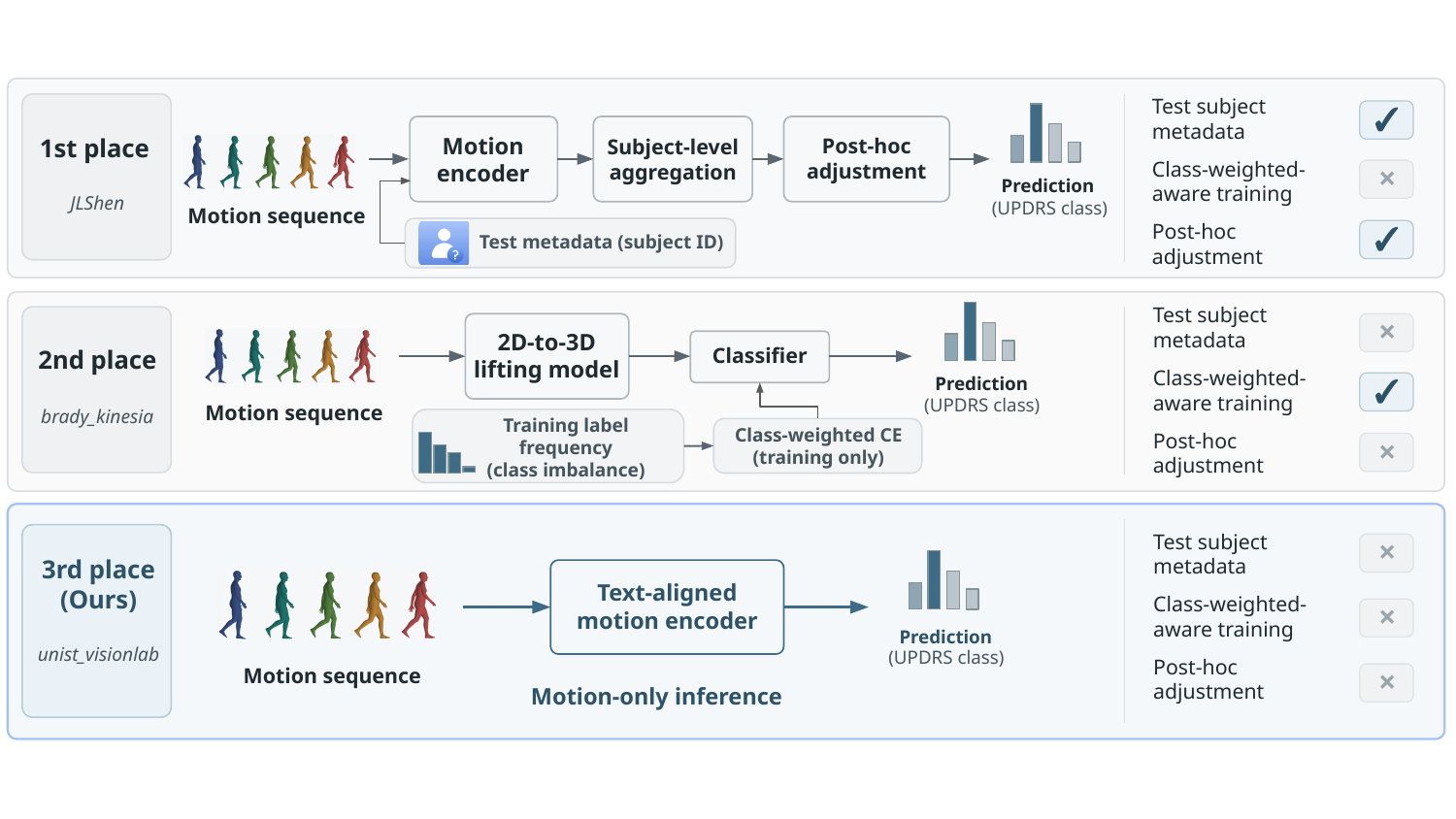}
    \caption{Comparison of information dependencies among the top-ranked MoCha challenge entries. Unlike the higher-ranked methods, our model does not rely on test-subject metadata, class-weighted training, or test-time adjustment, and predicts directly from motion observations alone. This setting imposes fewer assumptions on the unseen test domain and more closely reflects realistic deployment conditions.}
    \label{fig:intro}
\end{figure}

Human motion recognition in real-world and clinical environments is challenging because motion distributions can vary across subjects, cohorts, and acquisition sites. Such domain shifts can cause models trained on observed domains to rely on domain-specific patterns and consequently degrade when applied to unseen environments~\cite{Pitawela2025CLOC,Castro2020Causality,Su2025DomainAdaptive}. To address this issue, we investigate language-aligned motion representations that encourage the learned features to capture semantically meaningful gait characteristics while remaining transferable across heterogeneous clinical domains. The MoCha Challenge addresses this problem through cross-site evaluation on the CARE-PD dataset~\cite{Adeli}, which provides harmonized SMPL-based gait sequences collected across multiple cohorts and clinical centers.

We approach this problem from the perspective of fine-grained motion representation learning. Parkinsonian gait abnormalities often involve subtle temporal and kinematic differences, making it important to learn representations that capture discriminative motion patterns while remaining robust to domain-specific variations. To this end, we employ a Bi-GRU-based shared backbone to encode temporal dynamics from SMPL sequences. The backbone is optimized using classification and text-alignment objectives, allowing the learned embedding to capture both class-discriminative information and semantic relationships between motion classes. This is particularly relevant in the presence of class imbalance, where conventional supervised learning may become dominated by frequent classes.

After learning the shared representation, we separately adapt the backbone to each source domain. These domain-specific models can capture complementary motion characteristics associated with different cohorts, but relying on a single source model may limit generalization to unseen domains. We therefore perform parameter-level model merging~\cite{Chaves} to integrate knowledge from the independently adapted models. Rather than treating source domains as interchangeable samples from a single distribution, we regard them as complementary sources of motion knowledge and aim to consolidate this knowledge into a unified model.

Beyond the proposed representation learning framework, our approach also differs from the higher-ranked challenge entries in the information required for prediction. The comparative pipelines in Figure.~\ref{fig:intro} highlight the key distinctions of our model. While the higher-ranked approaches rely on additional dataset-specific information or adjustment strategies, such as test-subject grouping, class-weighted training, or test-time calibration, our method operates under a more constrained information setting. In particular, we assume that such auxiliary information is unavailable and perform prediction directly from motion observations alone. This setting more closely reflects realistic deployment scenarios, where metadata, class-distribution priors, and test-set-level statistics may not be accessible for previously unseen subjects or clinical sites.

Thus, our overall framework consists of three stages: (1) shared motion representation learning, (2) source-domain-specific adaptation, and (3) parameter-level model merging. Through this formulation, we investigate whether combining complementary source-domain knowledge can produce a more transferable representation for fine-grained human motion recognition under the unseen-site setting of the MoCha Challenge.

\section{Methods}

\begin{figure}[!t]
    \centering
    \includegraphics[width=\linewidth]{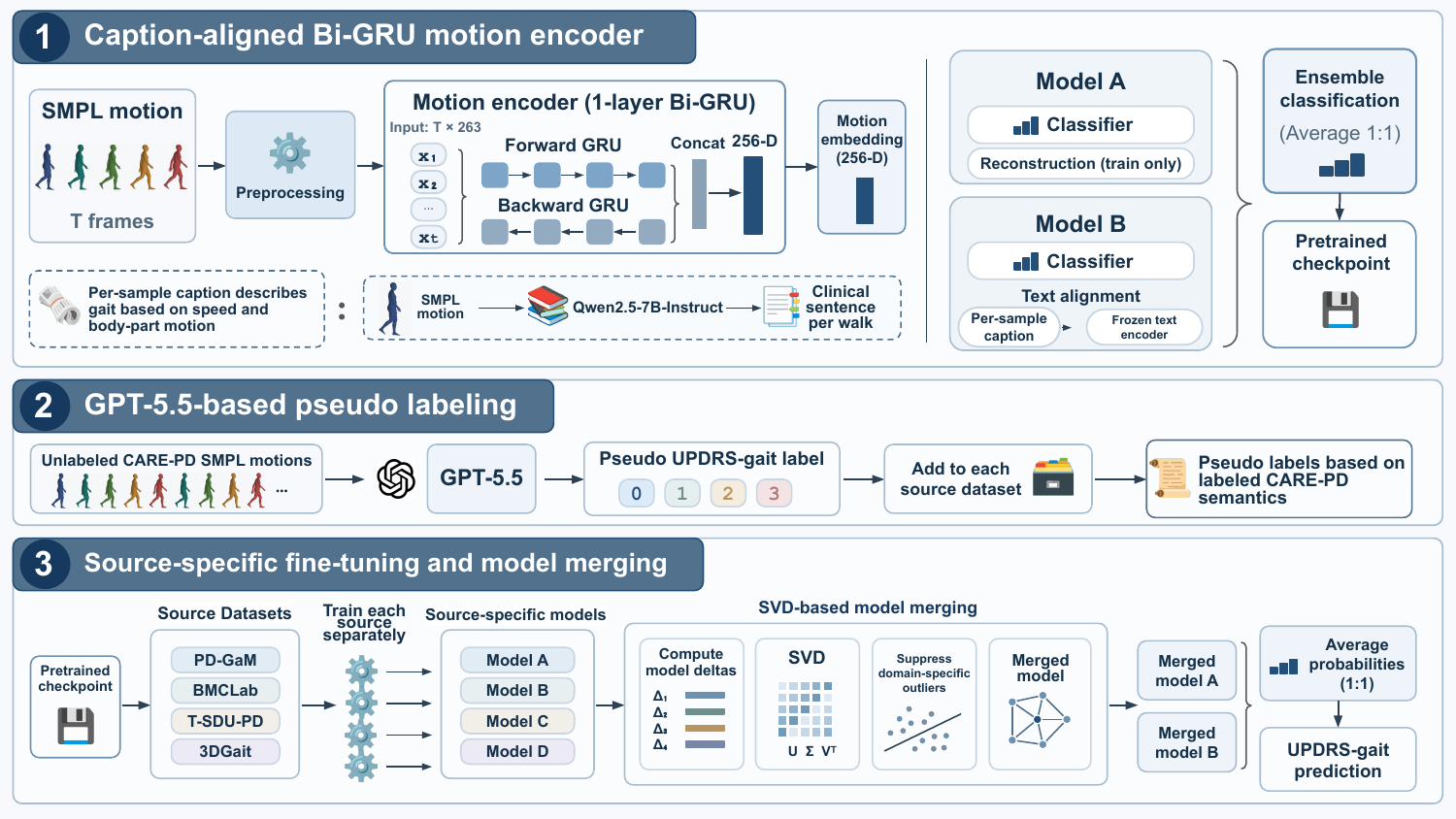}
    \caption{Overview of the proposed three-stage framework. (1) Bi-GRU motion representation learning trains two complementary branches with classification together with motion reconstruction or caption-based semantic alignment. (2) GPT-5.5-based pseudo labeling augments unlabeled CARE-PD motions with pseudo UPDRS-gait labels to mitigate class imbalance. (3) Source-specific fine-tuning and model merging independently adapts each pretrained branch to the source domains and consolidates the resulting models through SVD-based parameter merging. The predictions of the two merged branches are averaged to produce the final UPDRS-gait prediction.}
    \label{fig:method}
\end{figure}

Our framework Figure.~\ref{fig:method} consists of three stages: caption-aligned motion pretraining, GPT-5.5-based pseudo labeling, and source-specific fine-tuning followed by parameter-level model merging for domain generalization.

\subsection{Caption-aligned Bi-GRU motion encoder}

Each SMPL motion sequence is resampled to 25 FPS and converted into a 263-dimensional HumanML3D representation. Sequences are cropped or padded to 200 frames and normalized before being processed by a single-layer bidirectional GRU. The forward and backward hidden states are concatenated into a 256-dimensional sequence embedding, which is used for four-class UPDRS-gait classification.

To encourage the encoder to focus on clinically meaningful gait characteristics, we additionally align the motion embedding with textual gait descriptions. The captions describe severity-related motion patterns such as walking speed, stride and foot movement, arm swing, limb mobility, and postural stability. A frozen text encoder provides the caption representation, while a projection head maps the Bi-GRU embedding into the corresponding semantic space. A motion reconstruction decoder is also used only during training to preserve fine-grained temporal information.

Captions are generated offline and once. For each sequence we extract per-side knee flexion range, hip swing range and foot-lift height, their left--right asymmetries, cadence, step count and step-time variability, express each relative to its population tertile rather than as an absolute value, and pass the result to Qwen2.5-7B-Instruct~\cite{qwen2.5}, which writes one clinical sentence per walk (2{,}914 distinct captions from 2{,}915 sequences; no caption names a severity level).

\subsection{GPT-5.5-based pseudo labeling}

CARE-PD exhibits substantial class imbalance across UPDRS-gait severity levels. We therefore use additional unlabeled CARE-PD sequences by generating pseudo labels with GPT-5.5.

GPT-5.5 analyzes the SMPL motion and assigns an UPDRS-gait pseudo label based on motion characteristics learned from the labeled CARE-PD examples. The labeling process considers clinically relevant cues including walking speed, step and foot movement, arm swing, body posture, and overall gait restriction. The resulting pseudo-labeled samples are added to their original source datasets to reduce class imbalance, while generated motion descriptions are also used as additional text supervision when available.

\subsection{Source-specific fine-tuning and model merging}

Starting from the same pretrained Bi-GRU model, we independently fine-tune the network on the four CARE-PD source datasets: PD-GaM, BMCLab, T-SDU-PD, and 3DGait. This allows each model to adapt to the characteristics of its own clinical domain while maintaining a common initialization.

After source-specific training, we merge the four independently adapted models at the parameter level. For each source, we compute the parameter change from the common pretrained model and apply singular value decomposition to matrix-valued model deltas. The source updates are represented in a shared basis, where unusually large domain-specific components are suppressed before reconstructing the merged parameters. This procedure is designed to preserve parameter changes that are consistently useful across domains while reducing conflicting source-specific updates~\cite{Chaves}.

The two motion branches are merged independently. During inference, both merged models predict four-class UPDRS-gait probabilities, and their probability distributions are averaged to obtain the final prediction.

\section{Experiments}

\subsection{Experimental setup}
For each source dataset, namely PD-GaM, BMCLab, T-SDU-PD, and 3DGait, the pretrained weights of Model A and Model B were used as identical initializations. Each source model was then independently fine-tuned for 15 epochs using both ground-truth and pseudo-labeled samples with equal sample weights. We used AdamW optimization with a weight decay of $1\times10^{-4}$. The learning rate was set to $2\times10^{-6}$ for the Bi-GRU backbone, $7\times10^{-5}$ for the classification and projection heads, and $1\times10^{-4}$ for the training-only decoder and CLIP adapter. The batch size was set to 96 and the random seed to 42. A cosine annealing learning-rate scheduler and gradient clipping with a maximum norm of 5.0 were applied during training.

We did not use validation-based best-checkpoint selection or early stopping. Instead, the final checkpoint from the 15th epoch was retained for each source domain. The resulting source-specific models were merged using SCORE~\cite{Chaves} with $\tau=1.96$ and a merge scale of 1.0. During final inference, the class probabilities predicted by the merged Model A and Model B were averaged with equal weights of 1:1. No post-hoc logit adjustment or class-prior correction was applied.

\subsection{Results}

\begin{table*}[t]
\centering
\caption{Performance comparison on the challenge leaderboard. Our entry achieved 3rd place, alongside results from the top 5 entries. Metrics include Macro-F1, Macro-Precision, Macro-Recall, Accuracy, and Quadratic Weighted Kappa (QWK).}
\begin{tabularx}{\textwidth}{@{}lYYYYY@{}}
\hline
\multicolumn{1}{c}{Participants}
& F1 & Precision & Recall & Acc & QWK \\
\hline

1st place (JLShen)~\cite{Shen}
& 0.69 & 0.72 & 0.68 & 0.66 & 0.60 \\

2nd place (brady\_kinesia)~\cite{Caiola}
& 0.58 & 0.65 & 0.55 & 0.53 & 0.41 \\

\rowcolor{blue!15}
\textbf{3rd place (unist\_visionlab)}
& \textbf{0.57}
& \textbf{0.56}
& \textbf{0.59}
& \textbf{0.54}
& \textbf{0.43} \\

4th place (Nottingham\_RVCE)
& 0.56 & 0.60 & 0.53 & 0.54 & 0.42 \\

5th place (tuananh1007)
& 0.55 & 0.60 & 0.52 & 0.49 & 0.43 \\

\hline
\end{tabularx}
\label{tab:leaderboard}
\end{table*}

We achieved 3rd place overall, and the official leaderboard results for the top five teams are reported in Table.~\ref{tab:leaderboard}.

\begin{table*}[htbp]
\centering
\caption{Comparison of benchmark-specific information and adjustment strategies used by the final top-ranked challenge entries. The winning entry uses the released anonymized subject grouping for subject-level posterior aggregation and applies label-free transductive adjustment using statistics of the hidden test set~\cite{Shen}. The runner-up trains its severity classifier using class-weighted cross-entropy~\cite{Caiola}. In contrast, our final model uses neither the released test-subject grouping nor test-time adjustment, and does not employ class-weighted training. This places our method in a more constrained information setting, where auxiliary metadata and test-set-level statistics are assumed to be unavailable, better reflecting deployment scenarios in which predictions must be made directly from motion observations alone. The row highlighted in blue corresponds to our model.}
\begin{tabularx}{\textwidth}{@{}lYYY@{}}
\hline
\multicolumn{1}{c}{Method}
& Subject grouping
& Class-weighted
& Test-time adj \\ 
\hline

1st place (JLShen)~\cite{Shen}
& \ding{51}
& \ding{55}
& \ding{51} \\

2nd place (brady\_kinesia)~\cite{Caiola}
& \ding{55}
& \ding{51}
& \ding{55} \\

\rowcolor{blue!15}
\textbf{3rd place (unist\_visionlab)}
& \ding{55}
& \ding{55}
& \ding{55} \\

\hline
\end{tabularx}
\label{tab:intro}
\end{table*}

In Table.~\ref{tab:intro}, the winning entry~\cite{Shen} exploits the anonymized subject grouping provided with the hidden test set for subject-level posterior aggregation and further applies label-free transductive adjustment using unlabeled test-set statistics. In contrast, the runner-up~\cite{Caiola} does not rely on test-set information, but trains its severity classifier using class-weighted cross-entropy. Our final model uses neither test-subject grouping nor test-time adjustment and does not employ class-weighted training. At inference, predictions are obtained directly from the input motion sequence, without auxiliary information from the unseen test cohort.

\begin{table*}[htbp]
\centering
\caption{
Challenge hidden-site evaluation.
F1-score, precision, and recall are reported as macro-averaged metrics
across the four UPDRS-gait classes.
We additionally compare the effects of pseudo-labeled data,
domain-specific model merging, and post-hoc logit adjustment
on the final challenge performance.
}
\begin{tabularx}{\textwidth}{@{}c|YYc|YYYYY@{}}
\hline

& & & &
\multicolumn{5}{c}{Hidden-site evaluation}
\\
\cline{5-9}

Methods
& Pseudo
& Merging
& \mbox{Logit-adj}
& F1
& Precision
& Recall
& Acc
& QWK
\\
\hline

Simple ReCon
& \ding{51}
& \ding{55}
& \ding{55}
& 0.53
& 0.54
& 0.53
& 0.53
& 0.37
\\

Simple ReCon
& \ding{55}
& \ding{55}
& \ding{51}
& 0.54
& \textbf{0.60}
& 0.51
& 0.54
& 0.41
\\

Simple ReCon
& \ding{51}
& \ding{51}
& \ding{55}
& 0.53
& 0.56
& 0.53
& \textbf{0.55}
& 0.38
\\
\hline

Backbone
& \ding{55}
& \ding{55}
& \ding{51}
& \textbf{0.57}
& \textbf{0.60}
& 0.55
& 0.52
& 0.41
\\

\rowcolor{blue!15}
\textbf{Ours}
& \ding{51}
& \ding{51}
& \ding{55}
& \textbf{0.57}
& 0.56
& \textbf{0.59}
& 0.54
& \textbf{0.43}
\\
\hline

\end{tabularx}
\label{tab:challenge_results}

\end{table*}

Table.~\ref{tab:challenge_results} compares the challenge performance of the pretrained backbone and our full framework. The pretrained model is trained only on the original ground-truth data, which exhibit substantial class imbalance. To compensate for this imbalance at inference time, we apply post-hoc logit adjustment~\cite{Menon} using the class prior estimated from the training set. In contrast, our full framework augments the training data through pseudo labeling, resulting in a substantially more balanced class distribution. Therefore, no post-hoc logit adjustment or class-prior correction is applied to our final model, and predictions are directly obtained from the merged model outputs.

The proposed model consists of two independent lightweight Bi-GRU branches, with approximately 637K parameters actively used during inference. In contrast, the Simple ReCon baseline, which adopts a simple encoder-decoder architecture, uses approximately 17.66M parameters along its inference path. Therefore, the proposed model performs prediction with approximately 27.7$\times$ fewer active parameters, reducing both model storage requirements and inference computation. Furthermore, in settings with limited training data and substantial distribution shifts across source domains, excessive model capacity may increase overfitting to source-specific characteristics. Thus, the lightweight architecture may also be beneficial for generalization.

\section{Conclusion}

In this work, we address UPDRS-gait classification under multi-domain distribution shifts and severe class imbalance. Our framework combines caption-aligned Bi-GRU motion pretraining, GPT-5.5-based pseudo labeling, and source-specific fine-tuning followed by parameter-level model merging. While the pretrained backbone requires post-hoc logit adjustment to compensate for the imbalanced ground-truth training set, the proposed framework directly performs inference without class-prior correction by constructing a more balanced training set through pseudo labeling. Overall, the proposed approach aims to learn more robust gait representations and reduce source-specific bias for improved generalization to unseen clinical domains.


%
%
\bibliographystyle{splncs04}
\bibliography{main}
\end{document}